\documentclass{Interspeech}
\interspeechcameraready

\usepackage[utf8]{inputenc}

\usepackage{cite}

\usepackage{svg}

\title{Investigating Transcription Normalization in the Faetar ASR Benchmark}

\author[affiliation={1,2}]{Leo}{Peckham}

\author[affiliation={1}]{Michael}{Ong}
\author[affiliation={1}]{Naomi}{Nagy}
\author[affiliation={1,2,3}]{Ewan}{Dunbar}

\affiliation{Department of Linguistics}{University of Toronto}{Canada}
\affiliation{Department of Computer Science}{University of Toronto}{Canada}
\affiliation{Department of French}{University of Toronto}{Canada}
\email{leo.peckham@mail.utoronto.ca, michael.ong@mail.utoronto.ca, naomi.nagy@utoronto.ca,
ewan.dunbar@utoronto.ca}

\begin{document}

\maketitle

\section{Context}

We provide a small but important update on the Faetar Speech Recognition Benchmark \cite{Ongetal2025}. The benchmark, initially released as a challenge task (with test data embargoed), is intended to teach us more about the domain of ``dirty'' low-resource ASR. 

We identified two major hurdles. First, due to an unfortunate error, one of the baselines for the constrained ASR task which interested most challenge participants had an incorrect phone error rate which was much lower than it should have been--the reported result in fact came from a different, unconstrained model.  We felt the impact of this as potential participants hesitated to submit when they were unable to beat this incorrect number.  This has since been corrected in the documentation. The archival link to \cite{Ongetal2025} in the references section here is correct, as is the leaderboard at \url{https://perceptimatic.github.io/faetarspeech/}. Any circulating papers listing the ESPnet reference system as constrained should be destroyed.

Second, we received many comments about the quality of the gold transcriptions themselves. As we ourselves emphasized, the only transcriptions we have are ``dirty'' in the sense that they contain various inconsistencies, some genuine (the same pronunciation is transcribed in different ways) and some simply because the data is by and large transcribed at a phonetic level, rather than a lexical or phonemic level like most corpora (the same word is not always pronounced in the same way, thus it is not transcribed in the same way). Furthermore, there is a higher-level inconsistency in the sense that transcription conventions may vary between being phonetic or phonemic/lexical. Importantly, this affected the test set, which means that there may be an ``irreducible error rate'' due to the effect of trying to predict something that is unpredictable given the acoustics. Some judged this to be a challenge too far. We have spent the intervening time reflecting on this issue and probing it further in the case of the Faetar benchmark. 

\section{Why transcribe?}

The issue of whether reference transcriptions are lexical, showing consistency whenever the same word is used, or phonetic, reflecting the pronunciation directly, points more broadly to a dual goal in automatic transcription, and in the transcription of spoken language more generally. This corresponds very roughly to the goal of a ``phone recognizer,''  outputting a sequence of phones, as distinct from that of ``automatic speech recognition,'' to output a sequence of words (although this distinction is not, in fact, so neat in the practice of automatic transcription, as we will see).

For linguistic research and documentation, both levels are essential. Without a detailed phonetic transcription, there is no way to systematically document and investigate pronunciation variation. However, linguistic research in general is not possible without a lexical-level transcription. This includes not only lexical, morphological, syntactic, and semantic research, but the study of pronunciation variation itself: such ``variation'' is predicated over the fact that the variability in question is not lexically distinctive, and thus it is imperative to correctly group together instances of the same word. Similar considerations are active among language learners and teachers, as well as in the related context of language documentation. The need is acute in the context of language revitalization, since minoritized languages do not tend to have extensive or accurate formal documentation, particularly when it comes to the range of variability \cite{BirdandKell}.

Learning to (automatically) transcribe at each level---phonetic or lexical---comes with its own challenges. The first challenge---the phonetic challenge---stems from the fact that accurately transcribing phonetic details may be more difficult as it requires more fine-grained acoustic distinctions to be made. On the other hand, the second challenge---the abstractness challenge---comes from the fact that learning to directly map acoustics to abstract phonemic or orthographic forms introduces the challenge of mapping disparate acoustic forms to a single symbol, in addition to the challenge of potentially modelling idiosyncrasies in the orthographic system.

The inconsistency of levels found in the Faetar transcriptions recalls in some ways the  ``eye dialect'' traditionally used in transcriptions of folklore,  occasional lapses into quasi-phonetic transcription. Although much of the criticism levelled at eye dialect does not apply to the Faetar corpus (most often often, that it is biased in a way that makes the speakers ``appear boorish, uneducated, rustic, gangsterish, and so on'' \cite{Preston1982}), the conflation of levels contributes to the irreducible error rate. Furthermore, it means that meeting the needs of a two-level transcription system---an important goal, even if it is not the stated immediate goal of the benchmark---would require ingenuity, to say the least.

\section{Two-level transcription and ASR}

Hybrid ASR systems with a fixed, word-based lexicon, can be used as two-level transcription systems. These systems, at least as traditionally constructed, are limited in two important ways. First, although pronunciation lexicons do allow for multiple variants to be associated with a single headword, these lexicons typically have a finite list. Even when tools have been used to explicitly model pronunciation variability, typically the goal is to have the variability ``compiled out'' to a finite lexicon, rather than to  use the generative mapping directly during decoding (for example, \cite{choi2025data}). Unexpected variability cannot be guaranteed to be accurately transcribed. Second, finite word-based lexicons suffer from out-of-vocabulary issues. This is typically resolved by backing off to a partially or wholly subword based lexicon, but, at that point, the two-level transcriber property either disappears---instead of trying to assign phonetic transcriptions to subwords, the acoustic model is made to predict characters directly---or the lexicon ceases to properly account for pronunciation variability, as generic grapheme-to-phoneme models are used to generate pronunciations for subwords \cite{wang2020investigation}.

More important than all this, the phonetic variation in most ASR lexicons is limited to major, well-documented variants which are phonemically distinct. This is in part because most manual ``phonetic'' transcriptions found in corpora are still mostly phonemic and lack very fine detail (as in TIMIT \cite{TIMIT}), and in (larger) part because the majority of pronunciation lexicons are not generated from manual transcriptions at all, but, rather, are static, pre-existing resources which do not reflect any particular corpus. True phone recognition is a marginal task in practice. This in part reflects the (perceived) target user group: fully literate speakers of a language, with a standardized orthography, who have no particular need, pedagogical or otherwise, to understand the range of pronunciation variation in the language (and, even if they did, it would already be well-documented elsewhere anyway).

Instead, the main perceived value of pronunciation lexicons is that they allow for top-down information to be leveraged during decoding. Lexical-level transcriptions, and word tokens in particular,  provide a more useful vocabulary than fine-grained phonetic symbols  for making predictions on the basis of syntactic and semantic expectations (they add ``textual predictability'' in the sense of \cite{Robertson2024}; see also \cite{fourtassi2014self}). Acoustic models, on the other hand, are orders of magnitude less complex if they can be made to focus on the task of transcribing individual phones, rather than whole words \cite{jelinek1975design}. Pronunciation lexicons merely provide the link between these two levels of temporal granularity. Leveraging this fact only---that text provides more direct access to higher-order linguistic structure which helps smooth decoding using top-down information---is not crucially dependent on the pronunciation lexicon being at any particular level of transcription detail.

More recently, explicit, two-level pronunciation lexicons have given way to end-to-end systems \cite{graves2006connectionist}. These  directly predict orthography, today typically in the form of subwords, without passing through any explicitly trained intermediate level. End-to-end models have been used fruitfully without any kind of extrinsic textual language model, under the assumption that large models, provided with enough data, can implicitly learn to exploit textual predictability---although more recent trends toward fusion with large language models \cite{ma2024embarrassingly} tilt back toward the use of explicit textual data in training. 

In low resource settings, end-to-end speech recognition poses several challenges, over and above the fact that it does not leave room for the two-level transcriptions that may be of more interest in these settings. First, it assumes there is a standard orthography: the ``end'' for the end-user is assumed to be present in the training data. In languages like Faetar, where there is no standard orthography, normalization of the transcriptions can only come via some additional extrinsic resource, or by automatically learning a lexicon or a general mapping from unnormalized transcriptions to lexical forms. Second, even if we had parallel lexical-level transcriptions, the end-to-end setting imposes the  challenge of abstractness. In a low resource setting, the phonetic challenge may be more feasible, and, for this reason, two-level models may be more appropriate. Finally, learning textual dependencies implicitly as part of the process of ASR training, as opposed to directly from text as part of an external language model, may not work well with smaller amounts of data. For this reason, hybrid models may be a better solution.

\section{Objectives}

\begin{figure}[th]
    \caption{Histogram of counts of variants from the test lexicon}
    \label{fig:variants}
    \centering
    \includegraphics[width=\linewidth]{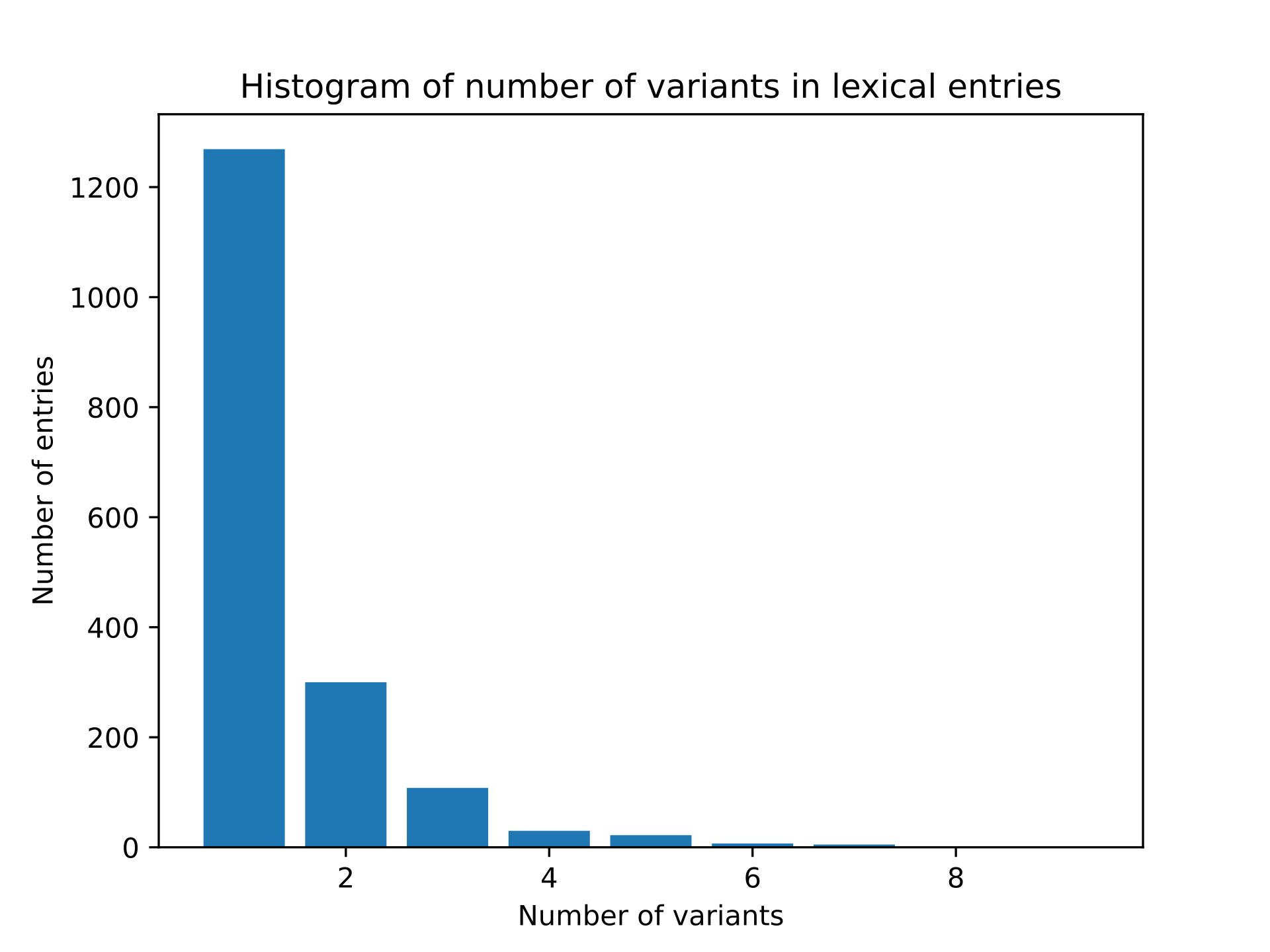}
\end{figure}

With these considerations in mind, we sought to evaluate whether or not the Faetar benchmark task of building a phone recognizer with (and evaluating against) dirty phonetic transcriptions, with no lexicon, and with no additional textual data for language modelling, is a useful and representative task as is---or whether the data or the task would need to be changed in order to make and to measure further progress towards a phone recognizer (and thus eventually toward a two-level transcription system). To facilitate this, we first collected an expert-curated lexicon for the test set and for the 1hr training subset. The lexicon groups together the attested tokens into words. Headwords are chosen arbitrarily. In the test subset, there are 9119 total words, 2509 of which are unique. The test lexicon groups these together into 1742 unique variants, of which 1269 are singletons. In the 1h subset, there are 10445 total words, 2769 of which are unique. The 1h lexicon groups these into 1910 unique variants, of which 1414 are singletons. A plot of the test set's variants is given in figure \ref{fig:variants}. Additionally, there is limited overlap between the variants in both sets, with only 754 variants being common to both of them.

Our previously reported phone error rates in \cite{Ongetal2025} were 30.5 for unconstrained systems, and 56.7 for constrained systems not using external resources. We first sought to establish whether the ``dirty phonetic challenge'' posed by training on, and evaluating against, transcriptions that are both phonetically fine-grained, capturing sub-lexical variation, and also somewhat internally inconsistent, was introducing an unreasonable challenge that might be preventing any further improvement.

One way of addressing this is to convert the reference transcriptions to headwords with the use of our lexicon. This does away with both real phonetic, sub-lexical variability, and variability introduced by inconsistencies in the transcription. It corresponds more or less to the approach taken by end-to-end systems. However, it introduces a challenge of abstractness. If the resulting task is easier, in the sense that these normalized transcriptions allow us to obtain lower phone error rates or word error rates, then we conclude that it would be worth exploring further normalization, either by developing more lexical resources or by explicitly considering some form of lexicon discovery as a part of the benchmark task. The challenge of abstractness thus introduced may outweigh the (dirty) phonetic challenge. In this case, the resulting task may actually be harder, given the limited amounts of data we have. Even if inconsistencies in the transcriptions are a problem, the situation could  not be easily improved using the straightforward strategies one might apply in high-resource languages like English.

The dirty phonetic challenge is of particular interest with the advent of universal phone recognition systems such as Allosaurus \cite{Allosaurus}. In principle, the output of these systems can be used as seed data to bootstrap a language-specific phone recognizer when transcriptions are not available \cite{bartels2016toward}. However, the output of these systems is extremely noisy. Evidence that cleaning up the inconsistencies in the Faetar transcriptions (which are quite minor in comparison to the errors  in current universal phone recognizers) by mapping to a lexicon leads to more consistent phone recognition in small datasets would be extremely significant---as would evidence to the contrary. To the extent that tokens from large pre-trained speech or audio models can be seen as noisy and/or fine-grained transcriptions, the same line of reasoning applies. 

Having access to a small lexicon also allows us to perform additional explorations. Top-down information about syntactic and semantic dependencies can be integrated explicitly in word-based hybrid systems, to the benefit not only of word error rate, but of phone recognition as well. However, these dependencies are obscured when sub-lexical transcriptions are used. Having a lexicon allows us to build a word-based hybrid system. Of course, there is no doubt that top-down lexical information can improve decoding. We are interested not in the general question but in the Faetar benchmark task and in settings like it. Are the (very) small amounts of transcribed data we have beneficial for bringing top-down information to bear on phone recognition?

\section{Results}

For our experimental data, we used the \emph{test} and \emph{1hr} splits as described in \cite{Ongetal2025}. We used a naïve iterated clustering algorithm based on Word2Vec \cite{Word2Vec} vectors concatenated to Hubert \cite{Hubert} vectors to generate a tentative lexicon. This lexicon was then sent to an expert for manual editing. This was done twice, once for the \emph{test} lexicon, and once for the  \emph{1hr} lexicon. We then preprocessed our transcription files, converting any word found in a test transcription file with its class representation (headword) as defined in the lexicon file.


As our model, we used the MMS-1B with continued pretraining and finetuning, as described in \cite{Ongetal2025}. This was one of our higher-performing models. For experiments using a language model, we used a bi-gram word-level modified Kneser-Ney language model. We used an $\alpha^{-1}$ of 1. We trained these language models  on \emph{test}  to get as close to an oracle as possible. Because we used \emph{test} to train our LM, in our experiments using language models, we refer to \emph{1hr} as \emph{AM train} and \emph{test} as \emph{LM train}.

\begin{table}[th]
    \caption{ASR without an LM, and varying levels of normalization in the data}
    \label{tab:exp1}
    \centering
    \begin{tabular}{ l l r r r r }
        \toprule
        Finetuning & Evaluation & \multicolumn{2}{c}{PER} & \multicolumn{1}{c}{WER} \\
        \multicolumn{1}{c}{Data} & \multicolumn{1}{c}{Data} & \multicolumn{1}{c}{Train (1h)} & \multicolumn{1}{c}{Test} & \multicolumn{1}{c}{Test} \\
        \midrule
        Original & Original     & 31.6 & 31.6 & 79.1 \\
        Normalized & Normalized & 35.1 & 36.2 & 82.3 \\
        Original & Normalized   & 38.1 & 33.8 & 80.4 \\
        \bottomrule
    \end{tabular}
\end{table}

We first evaluated whether using normalized transcriptions for training and test data made the phone recognition task easier or harder. The results are shown in Table \ref{tab:exp1} and show that, in fact, the task is harder, not easier. The problem of abstractness outstrips the problem of phonetic transcription.  We also evaluated how a model trained on the original transcriptions and evaluated against the normalized data would perform. The result, surprisingly, is actually better than when the system is trained on normalized data in terms of PER. Thus, at least part of the difficulty arising out from the use of normalized transcriptions must come from the challenge of training. 

\begin{table}[h]
    \caption{ASR trained on phonetic data using a bi-gram (lexical) word-based language model on phonetic data, extracting phone-level hypotheses}
    \label{tab:exp2}
    \centering
    \begin{tabular}{ l c c c }
        \toprule
        & \multicolumn{2}{c}{PER} & \multicolumn{1}{c}{WER} \\
        \multicolumn{1}{c}{LM} & \multicolumn{1}{c}{LM Train} & \multicolumn{1}{c}{AM Train} & \multicolumn{1}{c}{LM Train} \\
        \midrule
        Bi-gram & 30.3 & 33.5 & 84.3 \\
        No LM & 31.6 & 31.5 & 80.4 \\
        Bi-gram/Reduced & 28.9 & 32.4 & 84.3 \\
        \bottomrule
    \end{tabular}
\end{table}

Table \ref{tab:exp2} shows results from using a language model. In the first two settings, we fine-tune on phonetic data; for the PER score, we use the phone-level decoding from the language model, against the original transcriptions, and for WER, we use the word-level decoding against the normalized transcriptions. In the third setting, we completely normalize all of our data, we fine-tune and evaluate on normalized data, and use a reduced lexicon with only one variant (the headword). We still use phone-level decoding for PER and word-level decoding for WER.

By comparing the first two settings from Table \ref{tab:exp2}, we try to evaluate whether top-down information could, in principle, improve upon the scores from the Original-Original setting in Table \ref{tab:exp1}. To do this, we used an oracle language model trained on  \emph{test} (LM train), which also helped us avoid  problems related to out-of-vocabulary words. We see from Table \ref{tab:exp2} that PER the LM train set improves when using a language model. This effect may simply come from using a finite lexicon---of particular help since 70\% of words in the LM train set are not present in the AM train set. This latter fact also helps explain why performance on the AM train set gets worse.

The WER, however, gets worse than if we directly compare the phonetic output of the phone recognizer (which, remembering that the WER is calculated with respect to the normalized words, is surprising: the phone recognizer is trained on the original transcriptions). What this suggests is that language modelling is not, in fact, helping to choose the right word---it is often leading to the wrong word being chosen. The improvement to PER would thus seem to be driven mostly by the constraining effect of having a finite lexicon. 

The reduced lexicon removes all variants but the headword, and the PER drops even further. This is somewhat surprising. In this setting it will often happen that the actually attested variant cannot be chosen at all, because it is not in the lexicon.  Although the variants are generally quite similar to each other, one would think that  the model would benefit from having the full, attested list of variants in the lexicon. However, although the headwords were chosen arbitrarily, they are in fact the most frequent variant in 80\% of all lexical entries with two or more variants. The more complex problem of choosing a variant apparently leads to the model too often choosing the wrong one.

\section{Conclusions}

Although there is certainly phonetic and transcription noise in the transcriptions in the Faetar ASR Benchmark, moving to normalized transcriptions (which remove this noise) actually makes the problem harder. This suggests that the noise in the transcriptions is not a major source of irreducible error. We speculate that the quality of the audio data itself plays a major role in the high PER. This experiment also confirms that end-to-end speech recognition is difficult in low-resource settings.

Using a two-level lexicon and an oracle language model improved PER, but only because it reduced the number of options for the decoding by introducing a finite lexicon. The words that the hybrid model chose were generally chosen poorly, and the pronunciation variants were generally chosen poorly. We do not claim to have done a rigorous test of the value of language modelling, and we hope that these problems can be addressed---perhaps with judicious use of the 19-hour unlabelled data set, or  with the use of external resources such as multilingual language models. However, the result does suggest that, in settings like this one, it pays to build simple models in which the inherent complexity of decoding is kept to a minimum.

\bibliography{library}

\bibliographystyle{IEEEtran}

\end{document}